# Convolutional Neural Network with Pruning Method for Handwritten Digit Recognition

Mengyu Chen

Research School of Computer Science,
Australian National University
u6013662@anu.edu.au

**Abstract.** CNN model is a popular method for imagery analysis, so it could be utilized to recognize handwritten digits based on MNIST datasets. For higher recognition accuracy, various CNN models with different fully connected layer sizes are exploited to figure out the relationship between the CNN fully connected layer size and the recognition accuracy. Inspired by previous pruning work, we performed pruning methods of distinctiveness on CNN models and compared the pruning performance with NN models. For better pruning performances on CNN, the effect of angle threshold on the pruning performance was explored. The evaluation results show that: for the fully connected layer size, there is a threshold, so that when the layer size increases, the recognition accuracy grows if the layer size smaller than the threshold, and falls if the layer size larger than the threshold; the performance of pruning performed on CNN is worse than on NN; as pruning angle threshold increases, the fully connected layer size and the recognition accuracy decreases. This paper also shows that for CNN models trained by the MNIST dataset, they are capable of handwritten digit recognition and achieve the highest recognition accuracy with fully connected layer size 400. In addition, for same dataset MNIST, CNN models work better than big, deep, simple NN models in a published paper.

## 1 Introduction

Handwriting recognition is the ability of a computer or device to recognize the input handwriting from various sources such as printed physical documents, pictures and other devices. Many techniques have been developed to recognize the handwriting, such as Optical Character Recognition (OCR), Support Vector Machine (SVM), K-Nearest Neighbor (KNN) and Neural Network [14]. The Convolutional Neural Network (CNN) models are widely exploited to deal with problems about images, such as face recognition [4], scene labeling [5], action recognition [6] and image classification [7]. MNIST [2] has a large number of handwritten digit images, providing sufficient training and testing instances for methods like machine learning [9], Limited Receptive Area (LIRA) [10], and Deep Boltzmann Machines (DBM) [11] to achieve hand-written digit recognition.

We want to explore the performance of CNN on handwritten digit recognition, so MNIST is utilized to provide training and testing datasets for CNN models. During experiments on CNN models, it is found that different sizes of the fully connected layer would result in different recognition accuracy. To explore the relationship between the size of the fully connected layer and the accuracy of handwritten digit recognition, further experiments are executed to show different recognition results caused by different fully connected layer sizes. Inspired by the previous pruning work on the simple neural network (NN), we want to find out whether the performance of pruning by sensitiveness [1] on CNN is as same as on NN. The experiments results show the pruning angle threshold used on NN models is not suitable for CNN models, so further experiments are executed to explore the relationship between pruning angle threshold and performance of pruning.

### 1.1 Problem Definition

The investigation aims are to explore the CNN performance for handwritten digit recognition with MNIST dataset, to explore the effects of the CNN fully connected layer size on the recognition accuracy, to compare the performance of pruning method by distinctiveness on CNN and NN models, as well as to explore the effects of pruning angle threshold on the pruning results.

### 1.2 Dataset

The MNIST Digit database (Modified National Institute of Standards and Technology database) [2], a large-scale database of handwritten digits, is commonly utilized for training image processing models, which can be downloaded from the website [2]. It contains a training set (60,000 grayscale images) and a testing (10,000 grayscale images) [10]. Each image contains 28 x 28 pixels, with a size-normalized digit on its center [2], and only belong to one of the 10 balanced classes. The label for each image is a number from 0 to 9, which is equal to the digit on its center.

The MNIST is chosen because it provides with a large scale of training handwritten digit instances, and the availability of a large quantity of training is a necessary factor for the success of CNN [4]. Besides, digits on images have been size-normalized, which means fewer works for data pre-processing are needed. In addition, 'distribution of the training data has a big impact on CNN performance, a balanced training set was optimal' [8]. The MINIST Digit provides a balanced training data, which benefits the performance of CNN.

### 1.3 Convolutional Neural Network Model

According to [16], convolutional neural network has an input layer, an output layer, as well as multiple hidden layers. Hidden layers contain convolutional layers, pooling layers, and fully connected layers. Convolutional layers extracts features from the input or previous layers by utilizing multiple filters. A pooling layer will then simply perform downsampling along the spatial dimensionality of the given input, which can also control overfitting. The output from convolutional layers and pooling layer are all modified by ReLU since it is much faster than other functions to achieve non-linear combinations. Before output layer, feed-forward fully connected layers are utilized to process high-level features. The input of fully connected layer would be added some noises by dropping out a part of data.

### 1.4 Pruning by Distinctiveness

[1] introduces a pruning method using distinctiveness of hidden units, which help to automatically locate redundant neurons.

Firstly, the normalized units output activation over the pattern set are needed. Secondly, calculate the angle between 2 vectors in pattern space: if the angle is less than or equal to 15°, these two hidden neurons are similar, so one of them need to be removed; if the angle is larger than or equal to 165°, these two hidden neurons are complementary, so all of them need to be removed. Weights also need to be adjusted when hidden neurons are removed, but retraining is not needed.

## 2 Method

Fig. 1 shows the process of experiments: MNIST data is loaded as training and testing sets to building CNN models. Next go into the training and evaluation iteration for the CNN model built in the previous step. Modification of connected layer size are needed to explore the effects of the CNN fully connected layer size on the recognition accuracy. The evaluation results for different hidden size are recorded. Pruning method would be performed on the best trained CNN model. Finally, use evaluation approaches to test the performance of pruning.

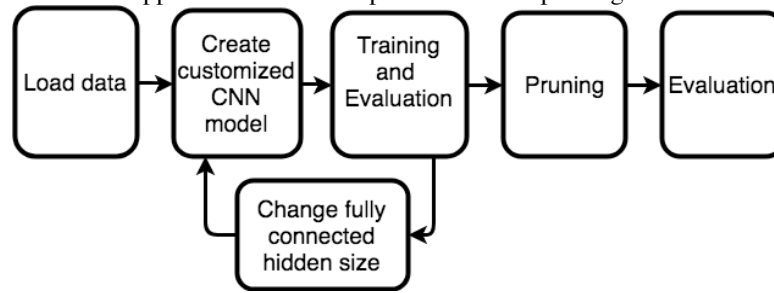

Fig 1. The whole process of experiments

### 2.1 Load Data

60,000 hand-written digit images from MNIST training dataset are loaded as the training set, while 10,000 hand-written digit images from the MNIST testing dataset are loaded as the testing set. Both the training set and the testing set have no overlapping instance. Images are normalized according to their mean value and standard deviation for better training and testing.

Using different ordering of training samples would possibly lead to a different local minimum for the objective function, so that different random shuffling of the training set would lead to different parameters [12]. It means that the performance of the CNN model would be affected by the order of training instances. To avoid CNN performance being devastated by extreme training instance order, 10 shuffled epochs are used to train CNN models. Before copied into each epoch, all the training instances are shuffled, which means each epoch has all the training data of MNIST, but the order of instances is different.

## 2.2 Create Customized CNN Model

As for input, each instance of the training set is an image, so each input for the CNN model is a 28 x 28 grayscale image. As for output, there are 10 classes of labels, so the output should be a vector with 10 elements showing the probability for each class. For the investigation aim that exploring the effects of fully connected layer size, the number of all the layers, the functionality of each layer and the size of all the other layers should be fixed except for fully connected layer. According to the input, the output, and the investigation aim, the CNN model is built.

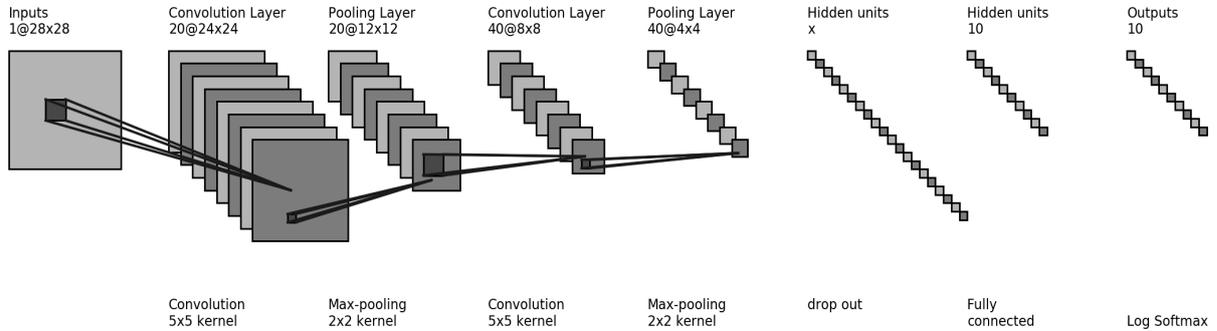

Fig 2. The CNN model

As shown in Fig. 2, two convolution layers are utilized following by a max-pooling layer for each. [13] shows that 5x5 kernel on the first layer can usually achieve the best results for MNIST images and for the max-pooling layer the typical size is 2x2. For the extraction of features, 20 filters are sufficient to get simple and general features and 40 filters are enough to get higher-level features. In experiments, the CNN models with 20 filters for the first convolutional layer and 40 filters for the second have higher recognition accuracy than 10, 20 for the two convolutional layers respectively. Meanwhile, their performances are very close to the CNN models with larger sizes. Therefore, 20 filters and 40 filters are used for the two convolutional layers. By dropping out, 25% data is removed randomly and the remaining data is passed to fully connected layer. The fully connected layers contain 2 hidden layers, but in this paper, the 'fully connected layer size' just refers to the size of one layer which is close to the convolutional layer, since the functionality of the other is to produce raw outputs later. The raw output should have the same size as the final output, so the size of the other fully connected layer (output layer) should be 10.

Log Softmax function is utilized to modify the raw output to a Log value, showing the probability of the input digit belongs to each of the classes from 0 to 9, which is required by the Negative Log Likelihood function to calculate the loss during the training period.

According to the rules-of-thumb, the fully connected layer size should be smaller than its previous layer size (640) and larger than its next layer size (10).

## 2.3 Training and Evaluation

For saving running time, both training and evaluation processes could be done by GPU.

**Training.** In training process, the training set is split into mini batches, with 64 hand-written digit images for each batch, because 64 is a commonly used batch size for CNN [15]. Since the whole training set has 60,000 images, it is both time-consuming and space-consuming to update the parameters using so large scale data. On the other hand, update parameters using every single instance would probably be noisy if some of the instances are not good representations of the whole dataset. So, we use mini-batch and set the batch size as 64, which can give a good representation of the whole data. For training CNN model, input images from the training set one by one to CNN model, and get a vector of 10 elements from the CNN output for the certain input. Compare with the output vector with target vector, for example, if the label of input is 3, the target vector should be [0,0,0,1,0,0,0,0,0,0] (with 10 elements, the elements with index 3 is 1 while others are 0), to get the loss for the certain input with loss function Negative Log Likelihood (NLL). Since the Softmax shows the probability of the input digit belonging to one of the classes, and this probability is between 0 and 1, when taking the log of that value, the value increases (and is negative), which is the opposite of what we want, so we simply negate the answer, hence the Negative Log Likelihood. That is also why Log Softmax is used to modify the final output. After getting the loss of the whole batch, calculate the gradient of the loss. Before calculation, the gradient of other batches should be cleared, otherwise, the new gradient of the current batch would be accumulated on other's gradients, which would give a wrong descent direction. Stochastic Gradient Descent (SGD) is used as update rules (optimizer) since it is suitable for mini-batch training, with learning rate 0.01 and momentum 0.05. Learning rate 0.01 and momentum 0.05 show a better recognition results when compared with other values. CNN parameters are updated by SGD rules and the parameters of the CNN are updated batch by batch until all the batches of the certain epoch have gone through the CNN. An evaluation would break up the training to test the current recognition ability after the training of each epoch. For a CNN model, the whole training process finished after been trained by all the 10 epochs.

**Evaluation.** When the training of a single epoch is finished, the evaluation process for the current CNN model starts. The input images for the evaluation are from testing set. The output vector of the CNN model is containing 10 elements, each one is a probability that the input digits belonging to 0-9. Find the index of the element with the maximum probability, which is the prediction or recognition class for the input handwritten digit given by CNN. Compare the prediction with the actual label of the input image, record the recognition result: whether the recognition is correct or not. After recognizing all the instances of testing set, make the number of correct recognition divided by the number of all the instances of the testing set to get the recognition accuracy for the current CNN model. After all the images of a single epoch have gone through the training process and evaluation, start the next iteration of training and evaluation for the next epoch, until all the 10 epochs run over.

For a CNN model, it would be evaluated for 10 times. The results of the first 9 evaluations during training is less important since it just reflects the performance of the immature CNN models but the 10th evaluation can indicate the final performance of the current CNN model. Therefore, only the result of the 10th evaluation is recorded and analyzed.

## 2.4 Change the Fully Connected Layer Size

In this step, only the size of the first fully connected layer (the order is from the input layer to the output layer) is changed within the range that smaller than the second pooling layer (640) and larger than the size of the output layer (10), to explore the effects of the fully connected layer to recognition accuracy. With the size of other layers, build a new CNN model initialed with small and random weights. Then repeat the last step: training and evaluation for the newly built model. Finally, record and compare the average and max accuracy of CNN models with different fully connected layer size.

## 2.5 Pruning

According to the distinctiveness in [1], and inspired by previous work, pruning could be done on CNN models as following steps shown in Fig. 3

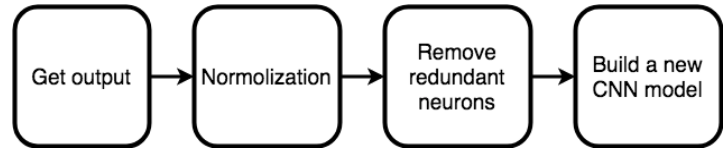

Fig 3. The process of pruning

**Get output.** For a CNN model, before ReLU process, get the raw output of the first fully connected layer (the one close to the convolutional layer). Since the angle between two output vectors would be calculated later and the range 0 to 1 of elements in output vector is required for the calculation of angle before normalization [1], the sigmoid function is used to modify the output to be between 0 and 1.

**Normalization.** Let the output vector of a hidden neuron in the fully connected layer be [v1, v2, v3, v4…, v10] and let every element in this vector subtract 0.5 to get [v1-0.5, v2-0.5, v3-0.5, v4-0.5 …, v10-0.5]

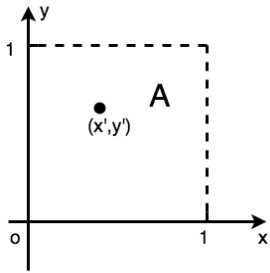 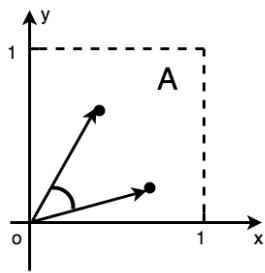 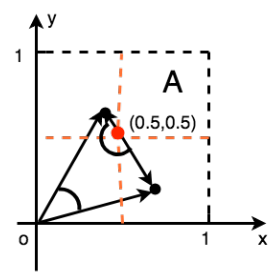

Fig 4. Points of vectors     Fig 5. Angles of vectors starting from (0,0)     Fig 6. Angles of vectors starting from (0.5,0.5)

The reason is that: for example, as shown in Fig. 4, for each 2D vector v = (x', y'), with 0< x'< 1, 0 < y'< 1, point of (x', y') must be in the square A. So, the angle between two vectors must between 0° and 90° (as shown in Fig. 5). By normalization, we move the starting point of vectors from (0,0) to (0.5, 0.5). In this case, the angle between two vectors is generally between 0° and 180° (as shown in Fig 6).

**Remove redundant neurons, change weights, bias of remaining neurons.**  For the angle $\alpha$ :

According to [1], to distinguish which pairs of hidden neurons are similar, and which pairs are complementary, we need to calculate the angle between two output vectors of the same pair. If the angle is less than $\alpha$, the pair of vectors is similar. In this case, we need to remove one of them and add the weights of left one to remaining one. If the angle is larger than 180°-$\alpha$, the pair of vectors is complementary. In this case, we need to remove both.

It's easier to get cosine value of angle than the exact angle degree. According to the value of cosine, when the angle is less than $\alpha$, the cosine value of this angle is larger than cos $\alpha$ and when larger than (180°-$\alpha$), the cosine value of this angle is smaller than cos(180°-$\alpha$)

Calculation the angle, after removing all the complementary pairs, remove one neuron for each the similar pair by adding the weights and bias of removed one to the remaining one.

The reason why similar pairs are removed later than complementary pairs is that, for example, there are hidden neurons h1, h2, and h3. h1 is similar to h2 with f1 (functionality of h1) = f2. Both h1 and h2 are complementary to h3 with f1 + f3 = 0 and f2 + f3 = 0. The integrated functionality of h1, h2 and h3 is f1+f2+f3 = f1 + (f2 + f3) = f1. If we deal with similar pairs firstly, we will remove h1 or h2 firstly, e.g. we remove h2 and left h1 and h3. When we deal with complementary pairs later, we will remove both h1 and h3. As a result, we removed the group of h1, h2, h3, which should have a contribution of f1 to the network. In this way, the performance of network must be affected. However, if we deal with complementary pairs firstly, e.g. remove h2 and h3, and make h1 left. The group of h1, h2, h3 still has a contribution of f1 to the network.

**Build a new neuron network with weights, bias of remaining neurons.**  Build a new network with the original size of convolutional layers, max-pooling layers, output layers and the new size of remaining hidden neurons in the fully connected layer. And make the weights and bias of the remaining items equal to the original one.

## 2.6    Evaluation for pruning.

With the new CNN model, test it with the testing set as evaluation in 2.3 does. Record and compare the accuracy rate and fully connected layer size for both the original CNN and the pruned CNN.

# 3    Results and discussion

## 3.1    Handwritten Digit Recognition Performance of CNN

Firstly, the effects of the fully connected layer size on the accuracy are tested.

The structure of each CNN model is as shown in Fig. 2, the only difference between them is the size of the first fully connected layer size. All the CNN models are trained on the MNIST Digits training database, and evaluated on the MNIST Digits testing database.

The average recognition accuracy for fully connected layer size 20, 40, 80, 120, 160, 200, 400, and 600 are tested and recorded respectively. All these tested sizes are smaller than the size of previous layer (640) and larger than the output layer (10). Note: for a single iteration of training and testing in section 2.3, 10 accuracy rates are produced. Only the accuracy rates for the last epoch is recorded.

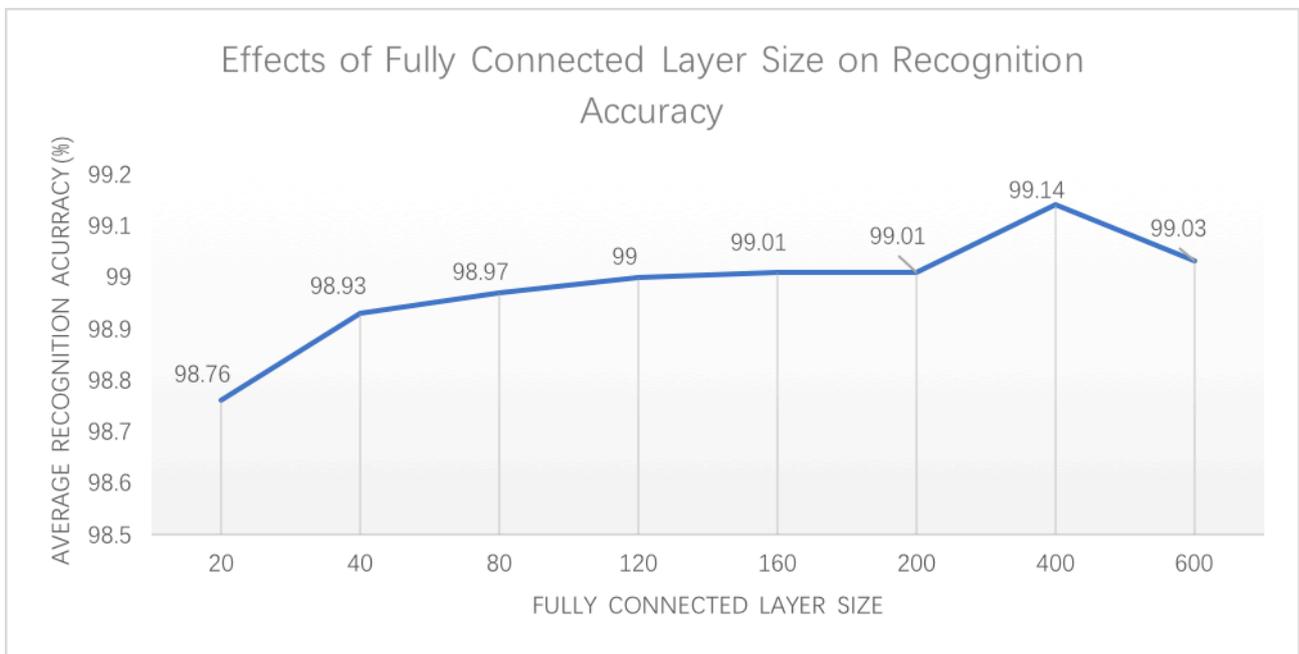

Fig 7. Effects of the fully connected layer size on recognition accuracy

Fig 7 indicates the average recognition accuracy with fully connected layer size from 20 to 600. With size 20, it has a minimum accuracy of 98.76%. From size 20 to size 40, there is an increase about 0.23% of recognition accuracy. From size 40 to size 200, the accuracy steadily grows as the size increases, with a slightly slower increasing speed than from size 20 to 40. Note the distribution of x-axis value is not uniform, although from the size 200 to size 400, this graph shows a sharp increment, the increment speed of accuracy are similar to the increment speed between 40 and 200. However, it still shows the highest accuracy rate 99.14% when with fully connected layer size 400. After 400, the accuracy starts dropping and falls to 99.03% with size 600, which is slightly higher than with size 200. From size 20 to size 600, the accuracies are all higher than 98.7%.

The results show that, although some CNN models have a small size of the fully connected layer, they can till recognize hand-written digits with sufficient high accuracies (higher than 98.7%), which means the high-level digit features extracted and modified by previous layers are good enough so that a small number of non-linear connection provided by neurons in fully connected layer can produce a good output. It also means the model of previous layers is very suitable for the MNIST Digits dataset. It can also be seen that the best fully connected layer size for the MNIST Digit for the CNN model created above is 400, which also means there is a fully connected layer size, when size increases but below than the threshold, the recognition accuracy rate keeps growing while when size is larger than the threshold, the recognition rate decreases as the size increases. It indicates that before the threshold size, the more non-linear connection there are, the more valuable the higher-level digit features produced by previous layers are further represented for correct recognition. However, when the fully connected layer is too large, the training hand-written digits would be over-fitted and had bad effects on recognition results. We can see from the graph, all the recognition accuracies are between 98.76% and 99.14% and the gap between the maximum and minimum value is just 0.38%, which means the effect of fully connected layer size on recognition accuracy is very small. Thus, for the CNN models that don't require extremely high accuracy on hand-written digit recognition, using a small size of fully connected layer could be considered to improve the recognition efficiency, but for the CNN models that require extremely high accuracy, more experiments should be implemented to find the best size which helps to achieve maximum recognition accuracy.

### 3.2 Performance of CNN Pruning with Different Angle Thresholds

In previous work, 15° is set as the pruning angle threshold for NN pruning of distinctiveness. After pruning, more than 25% of neurons are removed with a very little impact on testing accuracy. However, when 15° is set as pruning angle threshold for CNN, the percentage of removed neurons are very small. To achieve the better pruning performance, pruning methods of distinctiveness with different angle threshold are utilized to find better angle threshold for CNN pruning.

Since size 400 gives the best handwritten performance, 10 CNN models with 400 fully connected layer size were built, and trained with 10 epochs of MNIST Digits training set. Pruning was performed on the 10 CNN models respectively, with angle 10°, 15°, 20°, 25°, 30°, 35°, 40°, 45°, 50°, 55°, 60°, 65°, 70°, 75°, 80°. The remaining number of neurons of the fully connected layer and the recognition accuracy after pruning are recorded and shown in Fig. 8

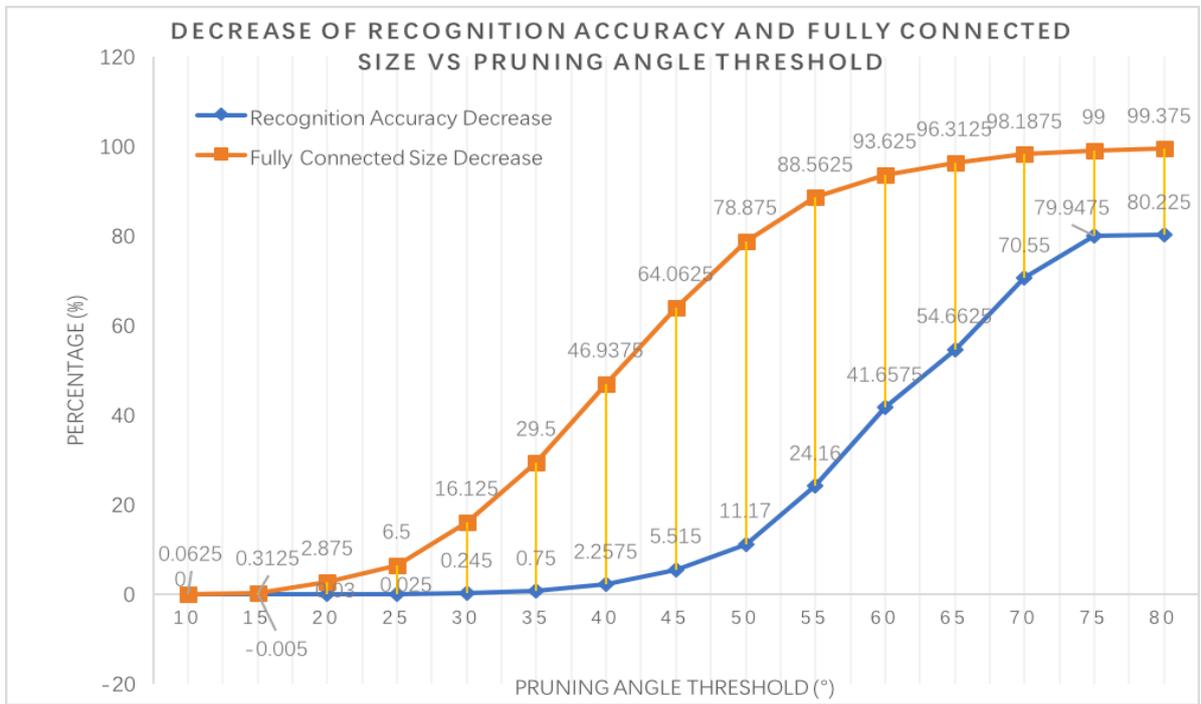

Fig 8. The decrease of recognition accuracy and fully connected size vs pruning angle threshold

Fig. 8 illustrates the changes of fully connected layer size and hand-written digit recognition accuracy as the pruning angle threshold. Here the pruning angle threshold means: for example, when the threshold is α (0° < α < 180°), in pruning steps, the two hidden neurons would be thought as similar if angle between their output vectors is less than α, and thought as complementary if the angle is larger than 180°-α. Note: the value of y-axis shows the decrease percentage, which means the more the y value is, the more decrease there is.

From Fig. 8, both fully connected size and recognition accuracy decrease as the pruning angle threshold grows. Since the increasing angle threshold would remove a larger proportion of neurons, it is obvious that the remaining fully connected size would decrease. With less of non-linear functionality provided by fully connected hidden neurons, less hand-written digits were recognized correctly. When angle threshold is less than 15°, the changes of both recognition accuracy and fully connected layer size are very tiny. That is because two hidden neurons of this kind have very similar output or quite complementary output, and removing didn't add too many noises for recognition. In paper [1], 15° is the best angle threshold for NN network, but for this CNN model, 15° can only remove 0.3125% hidden neurons on average. Although its accuracy is improved 0.0005% for controlling the overfitting, the percentage of removed neurons is much lower than in NN model. The reason is that CNN has a much larger training set than simple NN model, and after quite many iterations of parameters updates, the parameters between two different neurons has less probability to be similar or complementary. Thus, less pairs of vectors could have angle less than 15° or larger than 165°. As the angle rises from 20° to 35°, the layer size decreases gradually while the recognition keeps a high accuracy. It indicates that for this CNN model the pairs of neurons whose outputs angle are smaller than 35° are similar, and the pairs of neurons whose outputs angle are larger than 145° are complementary. 40° is another suitable angle threshold for pruning, with 2.2575% recognition accuracy decrease and nearly half of neurons removed. After 45°, with the steady decrease of fully connected layer size, the recognition accuracy drops sharply. When the angle is larger than 70°, although the fully connected size decreases to about 2%, the recognition accuracy is smaller than 30%, which is meaningless for handwritten digit recognition, and the number of remaining neurons also violate the Thumb-rule. From 10° to 50°, the gap between the decrease of layer size and recognition accuracy keep growing while after 50° the gap becomes more and more narrow, which means compared to the range from 50° to 80°, it decreases less recognition accuracy by removing more hidden neurons in range 10° to 50°. It can also be seen that from 20° to 65°, there is a dramatic drop for fully connected layer size from 2.875% to 96.3125%, which shows the 93.44% of outputs of neurons in the fully connected layer are in the range 20° to 65°.

There is a tradeoff between the fully connected layer size and the recognition accuracy. It would decrease recognition accuracy to make the layer size smaller. Therefore, for the CNN models which requires less accurate results, 40° or 45° could be the best threshold with less than 2.5% and 6% decrease of accuracy respectively. But for the CNN models which require quite accurate results, 30° or 35° could be the best threshold with less than 1% decrease of accuracy.

### 3.3 Comparison between NN Pruning and CNN Pruning

In previous work about NN pruning of distinctiveness, 15° is set as the pruning angle threshold. Since the database for NN is not MNIST, it's meaningless to compare the recognition accuracy of NN and CNN directly. However, the decrease of recognition accuracy and the decrease of hidden neuron size caused by pruning could be compared. The results of NN pruning are shown in Table 1.

Table 1. Performance comparison between simple neuron networks before and after pruning

| Number of hidden neurons | Average number of remaining hidden neurons | The percentage of removed hidden neurons (%) | Average testing accuracy before pruning (%) | Average testing accuracy after pruning (%) | Average improvement of testing accuracy (%) | Average change of testing time (seconds) |
|---|---|---|---|---|---|---|
| 20 | 14.9 | 25.5 | 92.25 | 92.07 | -0.180 | -0.0009 |
| 50 | 32.7 | 34.6 | 94.73 | 94.71 | -0.02 | -0.0024 |
| 100 | 57.7 | 42.3 | 95.03 | 94.98 | -0.060 | -0.0041 |
| 200 | 111 | 44.5 | 95.14 | 94.90 | -0.246 | -0.0067 |
| 400 | 198.6 | 50.35 | 95.15 | 95.04 | -0.106 | -0.0261 |
| 800 | 361.8 | 54.78 | 94.96 | 94.55 | -0.415 | -0.0390 |
| 1000 | 470.5 | 52.95 | 95.11 | 95.00 | -0.112 | -0.0601 |

Since the hidden neuron size of CNN for pruning is 400, we only compare the results with number of hidden neurons 400 in Table 1. For NN model, 50.35% hidden neurons are removed, which causes 0.106% decrease of testing accuracy. For CNN model, according to Fig.8, to remove 50.35% hidden neurons, the angle should be set around 41°, with 2.5% decrease on testing accuracy, which is much larger than 0.106% on NN; when decreasing 0.106% accuracy, CNN can only remove 27% neurons, which is fewer than 50.35% on NN.

The comparison indicates that for removing the same number of neurons, CNN would lose more testing accuracy than NN while for decreasing same percentage of testing accuracy, CNN would remove fewer neurons than NN. Therefore, the pruning of distinctiveness performs worse on CNN than on NN. Meanwhile, the suitable pruning angle thresholds for CNN are larger than 15°, which is the best pruning angle for NN [1].

The reason is that the higher-level features operated by the fully connected layer neurons in CNN are more complex than operated in NN, so that more neurons with unique significant functionality are required, which result in fewer redundant neurons in CNN fully connected layer than in NN hidden layer.

### 3.4 Comparison with Results of Related Paper

Paper [3] also use MNIST Digit database. Different from this paper, the model it used was deep, big, simple neural networks. However, its networks had many hidden layers, and a huge number of hidden neurons per layer. In addition to MNIST datasets, it also used graphics cards to greatly speed up learning.

Table 2. Testing results of NN models in the published paper

| ID | Architecture (Number of Neurons in Each Layer) | Test Error for Best Validation (%) | Best Test Error (%) | Simulation Time (h) | Weights (Millions) |
|---|---|---|---|---|---|
| 1 | 1000, 500, 10 | 0.49 | 0.44 | 23.4 | 1.34 |
| 2 | 1500, 1000, 500, 10 | 0.46 | 0.40 | 44.2 | 3.26 |
| 3 | 2000, 1500, 1000, 500, 10 | 0.41 | 0.39 | 66.7 | 6.69 |
| 4 | 2500, 2000, 1500, 1000, 500, 10 | 0.35 | 0.32 | 114.5 | 12.11 |
| 5 | 9 × 1000, 10 | 0.44 | 0.43 | 107.7 | 8.86 |

Table 3. Testing results of CNN models with fully connected layer size 400 in this paper

| Fully Connected Layer Size | Average Test Error (%) | Best Test Error (%) | Simulation Time (h) |
|---|---|---|---|
| 400 | 0.86 | 0.83 | 0.122 |

Table 2 shows the result in the paper [3]. The best test error is 0.32% with size 2500, 2000, 1500, 1000, 500, 10 hidden neurons respectively for each layer. In this paper, the best test error is 0.83% with fully connected size 400. The reason why [3] has a higher accuracy is that huge size of hidden neurons per layer can extract a huge number of features from the previous layer so that more common features of digits belong to the same class could be identified and learned. Besides, the larger number of layers provide more higher-level features, which helps to distinguish two different digits correctly. Compared with it, the CNN model in this paper has much smaller size, so that only a very small number of higher-level features are not extracted and learned.

However, although the accuracy rate of [3] is slightly higher than this paper, the simulation time in this paper is much less than that. Together with training and testing, the time consumed in this paper is just 0.122 hours for fully connected layer size 400 to get the best result, while 114.5 hours in [3] for highest accuracy. The time consumed in [3] is 939 times larger than in this paper.

In practice, it's unworthy to spend more than 100 hours to improve 0.51% accuracy on hand-written digit recognition. Thus, the result in this paper is better than [3].

## 4    Conclusion and Future Work

In conclusion, with MNIST Digit database, CNN in this paper can recognize the handwritten digits with 99.14% average accuracy. For the effects of the CNN fully connected layer size on the recognition accuracy, there is a threshold, when size increases but below than the threshold, the recognition accuracy rate keeps growing, in contract, when size is larger than the threshold, the recognition rate decreases as the size increases. However, when the fully connected layer size increases, the changes of recognition rate are very small. Thus, for the CNN models that don't require extremely high accuracy on hand-written digit recognition, using a small size of fully connected layer should be considered to improve efficiency. As pruning angle threshold increases, both the fully connected layer size and the recognition accuracy decreases. For the CNN models which requires less accurate results, 40° could be the best thresholds with less than 2.5% of accuracy drops. But for the CNN models which require quite accurate results, 35° could be the best threshold with less than 1% decrease of accuracy. Besides, the pruning angle threshold in CNN models should be larger than in NN models, within the range from 15° to 45°. In addition, most pairs of neurons in the fully connected layer has an angle between 20° and 65°.

Hand-written digits recognitions accuracy of CNN with smaller size is as similarly high as simple neural network with many layers and a huge number of hidden neurons per layer in a published research paper [3], but it runs much faster, so CNN works better than big, deep, simple NN models. It is found that suitable pruning angle thresholds for CNN are larger than for NN. When removing the same number of redundant neurons, CNN lose much more accuracy than on NN, so the performance of CNN pruning by distinctiveness is worse than NN.

The algorithm of pruning used in this paper can try to keep the functionality of whole network but cannot make sure the hidden layer be shrunken to a minimum size so it could be improved in the future to keep minimum number of hidden neurons left without changing the performance. In addition, from the experiment results, we found sometimes network even improved their accuracy rate after pruning, which could be researched in future to improve network performance. Besides, the removing ratio of the dropping out function before the fully connected layer is found has a significant influence on the pruning results. Further researches could focus on the effects of dropping out percentage for better pruning performance.